%% file: main.tex
\documentclass{article}
\usepackage{spconf,amsmath,graphicx}

\usepackage[utf8]{inputenc} 
\usepackage[T1]{fontenc}    
\usepackage{hyperref}       
\usepackage{url}            
\usepackage{booktabs}       
\usepackage{amsfonts}       
\usepackage{nicefrac}       
\usepackage{microtype}      

\usepackage{amssymb} 
\usepackage{color}
\usepackage{array}
\usepackage{mathtools}

\definecolor{red}{rgb}{0.65, 0, 0}
\definecolor{blue}{rgb}{0, 0, 1}
\definecolor{purple}{rgb}{0.5, 0, 1}
\definecolor{orange}{rgb}{ 0.84, 0.31, 0.078}

\newcommand*\conj[1]{\overline{#1}}

\renewcommand{\angle}{\measuredangle}

\def\imw#1#2{\fbox{\includegraphics[clip,width=#2\textwidth]{figures/#1.png}}}

\newcommand{\tb}[3]{\setlength{\tabcolsep}{#2mm}\begin{tabular}{#1}#3\end{tabular}}

\title{Better than Real: Complex-valued Neural Nets for MRI Fingerprinting}

\name{Patrick Virtue$^{\star \dagger}$ \qquad Stella X. Yu$^{\star \dagger}$ \qquad Michael Lustig$^{\star}$}
\address{$^{\star}$ University of California, Berkeley / $^{\dagger}$International Computer Science Institute}

\begin{document}
\maketitle

\begin{abstract}
\input{abstract}
\end{abstract}

\begin{keywords}
Magnetic Resonance Fingerprinting, Parameter Mapping, 
Complex-valued Neural Networks
\end{keywords}

\section{Introduction}
\label{sec:intro}

Fingerprinting in the magnetic resonance imaging (MRI) domain \cite{mrf_nature} quantifies tissue parameters from complex-valued MRI signals. 
Tissue in the body may be characterized by how it interacts with the magnetic field during a MRI scan. Two tissue parameters, T1 and T2, are exponential time constants, e.g. $e^{-t/\text{T}2}$, that describe how fast hydrogen protons in different tissues react to the applied magnetic field. For example, T1 and T2 values allow us to discern the boundary between gray matter (T1$=830$, T2$=80$) and white matter (T1$=500$, T2$=70$) in MRI brain images. These parameters also enable radiologists to differentiate between benign and malignant tissues.

\begin{figure}[t]
	\includegraphics[width=\linewidth,keepaspectratio]{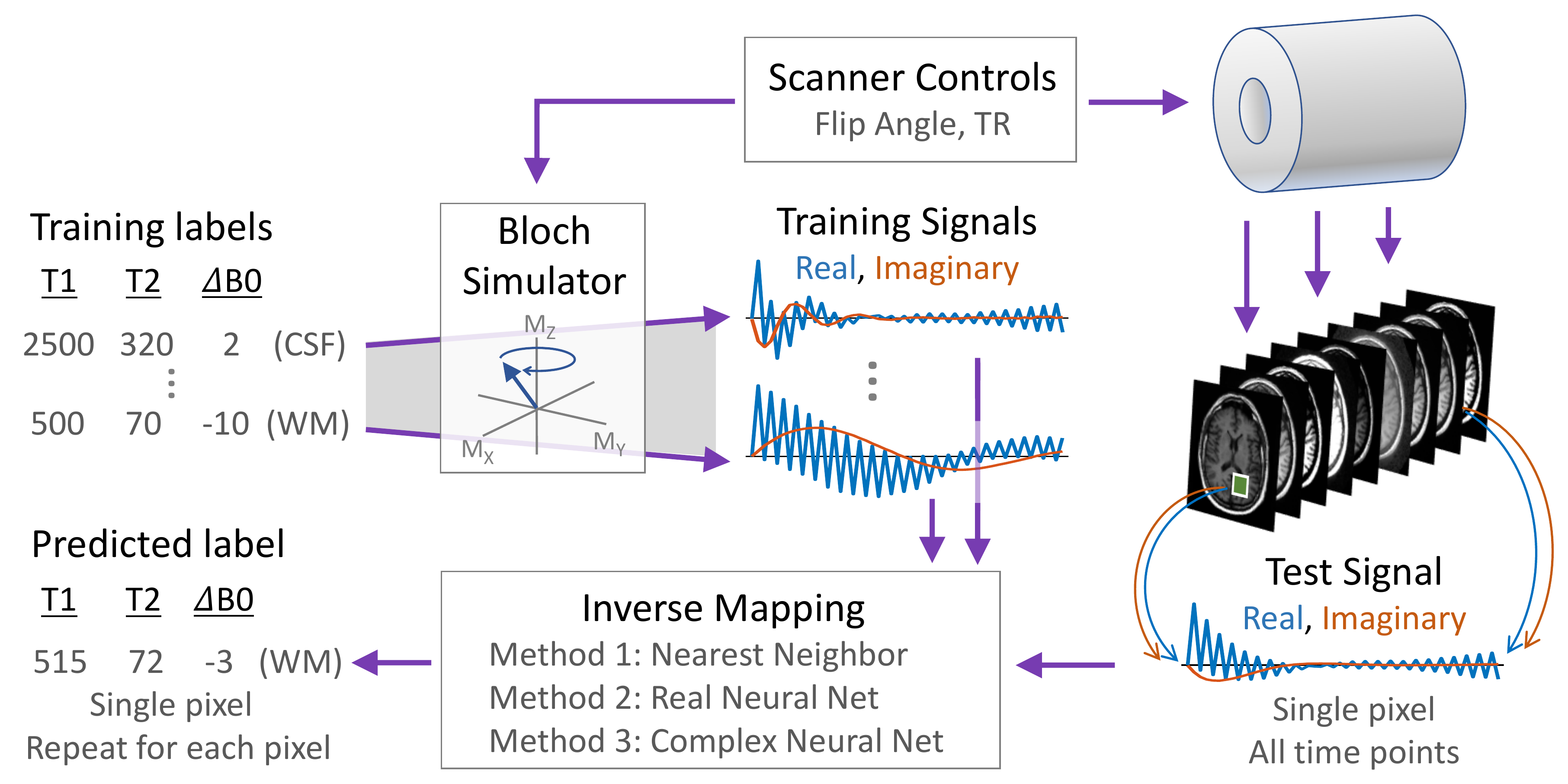}
	\label{fig:mrf}
	\caption{MRI fingerprinting is an inverse mapping problem that infers the tissue parameters from MRI signals. MRI simulator turns a ground-truth (T1, T2, B0) parameter tuple into an observed MRI temporal signal. The inverse mapping, by either nearest neighbor search or a neural network, solves for the tissue parameters given the simulated signal. At test time, the MRI signal will arrive from the scanner rather than the simulator.  Example complex-valued signals are shown for cerebral spinal fluid (CSF) and white matter (WM) parameters. }
\end{figure}

Traditional MRI requires {\it many different scans} that each accentuates one of the desired parameters. Additionally, those scans only provide a {\it qualitative} visual contrast between tissues, e.g. {\it in this image, tissues with high T1 are brighter than other tissues}. MRI fingerprinting as proposed in \cite{mrf_nature}, however, simultaneously produces {\it quantitative} values for T1, T2, and proton density in {\it one single scan}. It can also provide information about imperfections in the parameters for the applied  magnetic field, i.e. B0 and B1.

MRI fingerprinting works by scanning the subject using a predetermined progression of scanner controls, e.g. flip angles and repetition time (TR) values of the pulse sequence. The various tissues in the body will react to this pulse sequence, producing measurable signals that have unique signatures depending on their specific tissue parameter (T1, T2, proton density) and applied magnetic field parameters (B0, B1). Just like a fingerprint pattern can identify a specific person, these measured signals may be decoded to determine the tissue and magnetic field parameters at each pixel location in the image. Figure \ref{fig:mrf} shows how MRI fingerprinting uses a numerical simulator to convert parameters into MRI signals, which are then used to train an algorithm to solve the inverse problem of mapping the signal back to the original parameters. When scanning a patient, the unlabeled MRI signals will arrive from the scanner to be decoded by the inverse mapping algorithm.

From a machine learning perspective, the MRI simulator provides a potentially unlimited number of complex-valued MRI signals for training, each of these temporal sequences is paired with a tuple of real-valued labels (T1, T2, B0).  At the test time, the MRI scanner acquires a temporal signal at each pixel location, which must be decoded into to the tissue and MRI field parameters. Prior MRI fingerprinting works \cite{mrf_nature,mrf_multiscale,mrf_pssfp} have used a nearest neighbor search based approach to match the measured signal to a dictionary of simulated training signals. Due to the non-parametric nature of nearest neighbor methods, the computation time scales linearly with the size of the dictionary, quickly becoming infeasible with a finer parameter resolution or when more tissue parameters are required. Additional research has improved the nearest neighbor matching efficiency by incorporating SVD \cite{mrf_svd} and group matching \cite{mrf_groupmatching}.

Rather than non-parametric nearest neighbor-based methods, we propose to learn a parameterized model for solving the MRI fingerprinting inverse mapping problem. Specifically, we demonstrate that feedforward neural networks can accurately model the complex non-linear MRI fingerprinting inverse mapping function with a computational efficiency that does not scale with the number of training examples. 

We also investigate using complex-valued neural networks for MRI fingerprinting, since the MRI signals are inherently complex-valued. While complex-valued signals can be represented by 2-channel real signals, each channel containing real and imaginary components respectively, such a representation does not respect the phase information that is captured by complex algebra. 
Indeed, by introducing a new complex activation function for complex neural networks, we demonstrate that complex-valued neural nets are more effective than real-valued networks at MRI fingerprinting.

\section{Complex-Valued Neural Networks}
\label{sec:theory}

A significant facet of complex network research since the early 1990's has been to overcome the issue that standard real-valued non-linear layers do not transfer well to complex-valued networks.  We tackle several aspects here.

\subsection{Complex Cardioid Activation Function}

 Standard non-linear functions are either unbounded, e.g. $\text{sigmoid}(i\pi)$, or undefined, e.g. the max operator in max pooling and ReLU.  With complex outputs, we have also lost the probabilistic interpretations that functions like sigmoid and softmax provide. Past research has explored a range of potential solutions, for example, limiting the range of the activation input to avoid unbounded regions \cite{complex_backprop_leung}, or applying non-linearities to real and imaginary components separately \cite{complex_backprop_nitta,single_layer_cvnn}. In their 1992 paper \cite{complex_backprop_georgiou}, Georgiou and Koutsougeras presented an activation that attenuates the magnitude of the signal while preserving the phase. In our experiments, we refer to this activation function as \textbf{siglog} as it modifies the magnitude by applying the sigmoid of the log of the magnitude:
\begin{align}
\text{siglog}(z) &= \frac{z}{1+|z|} = g(\log(|z|))e^{-i\angle z}\\
g(z) &= 1/(1+e^{-z}).
\end{align}

\begin{figure}[tp]
\centering
	\includegraphics[width=0.28\linewidth,keepaspectratio]{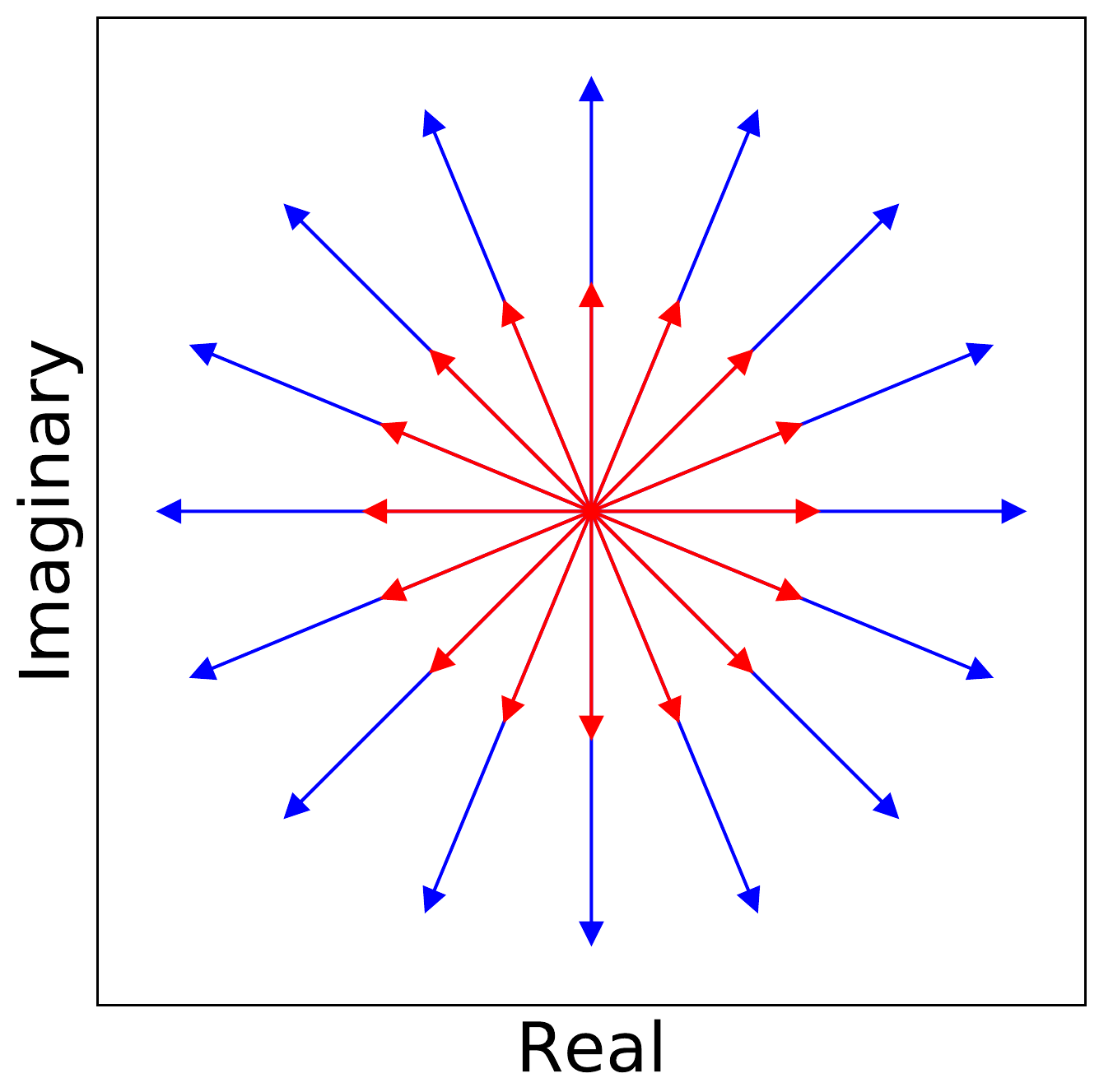}
	\includegraphics[width=0.28\linewidth,keepaspectratio]{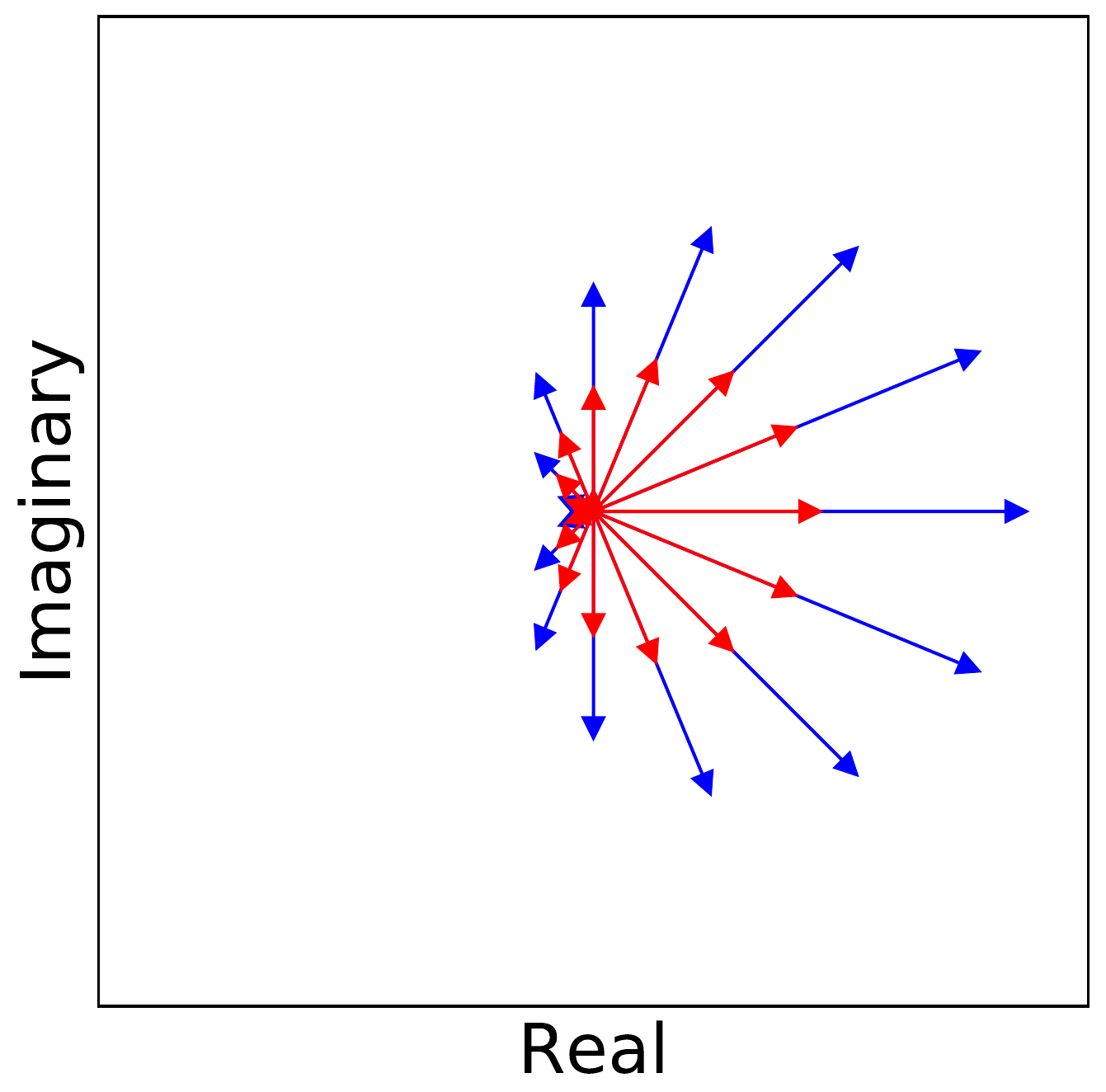}
	\includegraphics[width=0.40\linewidth,keepaspectratio]{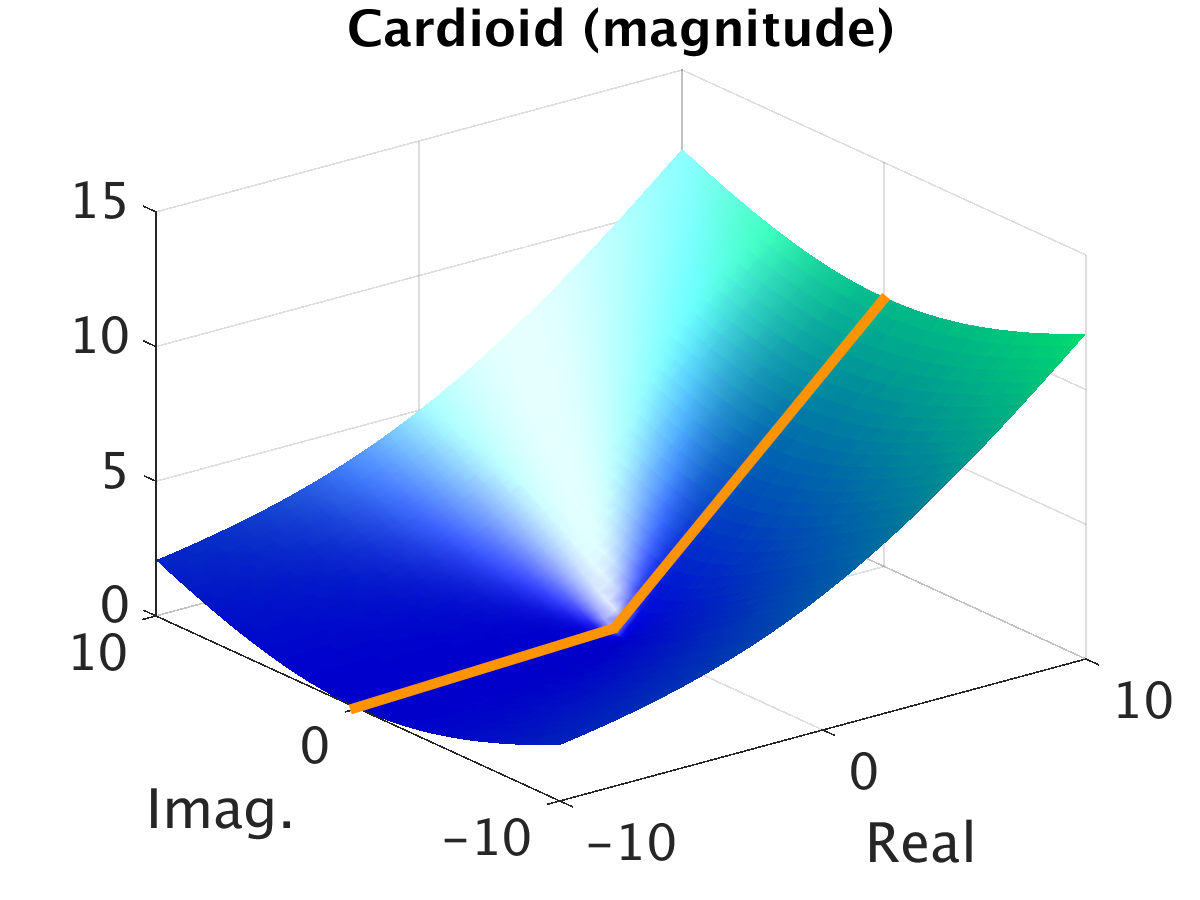}
	\label{fig:cardioid}
	\caption{Our new cardioid activation function is a phase sensitive complex extension of ReLU.  Left / Center: 
	Each arrow indicates a sample input/output of our cardioid function  on the complex plane. Right: The magnitude transformation of the cardioid function shows that it is reduced to ReLU on the real axis (orange line). }
\end{figure}

We propose a new complex activation function, \textbf{complex cardioid}, which is sensitive to the input phase rather than the input magnitude. The output magnitude is attenuated based on the input phase, while the output phase remains equal to the input phase. The complex cardioid is defined as:
\begin{align}
f(z)  = \frac12 ( 1 + \cos(\angle z) ) z
\end{align}
With this activation, input values that lie on the positive real axis are scaled by one, input values on the negative real axis are scaled by zero, and input values with nonzero imaginary components are gradually scaled from one to zero as the complex number rotates in phase from positive real axis towards the negative real axis. When the input values are restricted to real values, the complex cardioid function is simply the ReLU activation function.
The $\mathbb{CR}$ \cite{cr_calculus} derivatives are as follows:
\begin{align}
\frac{\partial f}{\partial z} &= \frac12 + \frac12 \cos\big(\angle z\big) + \frac{i}{4} \sin\big(\angle z\big) \\
{\color{red}\frac{\partial f}{\partial \conj{z}}} &= \frac{-i}{4} \sin\big(\angle z \big) \frac{z}{\conj{z}}
\end{align}

\subsection{Complex Calculus and Optimization}

We leverage Wirtinger calculus, or $\mathbb{CR}$ calculus, \cite{wirtinger,cr_calculus} to do gradient descent on functions that are not complex differentiable as long as they are differentiable with respect to their real and imaginary components. The first of two $\mathbb{CR}$ calculus derivatives is the $\mathbb{R}$-derivative (or real derivative) which computes $\partial f/\partial z$ by treating $z$ as a real variable and holding instances of $\conj{z}$ constant. Likewise, the other derivative is the {\color{red}conjugate $\mathbb{R}$-derivative}, {\color{red}$\partial f/\partial \conj{z}$}, where $\conj{z}$ acts as a real variable and $z$ is held constant.

To optimize a real-valued loss function ($L(w)$) at the end of a complex feedforward neural net,
we update the weights by applying the complex version of gradient descent:
\begin{align}
w &= w - \alpha \nabla_{\conj{w}}L
\text{, where } \nabla_{\conj{w}}L =
\bigg[{\color{red}\frac{\partial L}{\partial \conj{w}_1}} \hdots {\color{red}\frac{\partial L}{\partial \conj{w}_n}} \bigg] ^T
\end{align}
This is the same as real-valued gradient descent with careful attention paid to the gradient operator. As shown in \cite{brandwood}, the direction of steepest descent is the complex cogradient, $\nabla_{\conj{w}}L$.

\section{MRI Fingerprinting Experiments}
\label{sec:methods}

{\bf Training Data.} We simulate the MRI signal with the Bloch equations and the first of the two pulse sequence parameters from \cite{mrf_nature} with signal length 500. We use 100,000 simulated points for training, randomly sampled with the same T1, T2, B0 density as used in the baseline nearest neighbor dictionary.

{\bf Testing Data.} Following \cite{mrf_multiscale}, we test our methods with a numerical MRI phantom \cite{collins98,collins06} with the T1, T2, and proton density values specified for each tissue type in \cite{collins06}. We add a liner ramp in the B0 field across the image from -60 Hz to 60 Hz. We compute proton density from the norm of the test signal as in \cite{mrf_nature}. Although we do not include any B1 inhomogeneity in our experiments, a fourth neural network could easily be added to incorporate this or any other parameter(s).  In addition to testing with a clean signal from the phantom, we also tested phantom signals with complex-valued Gaussian noise added to produce a peak signal-to-noise ratio (pSNR) of 40.

\begin{figure}[bp]
\centering
	\includegraphics[width=0.9\linewidth,keepaspectratio]{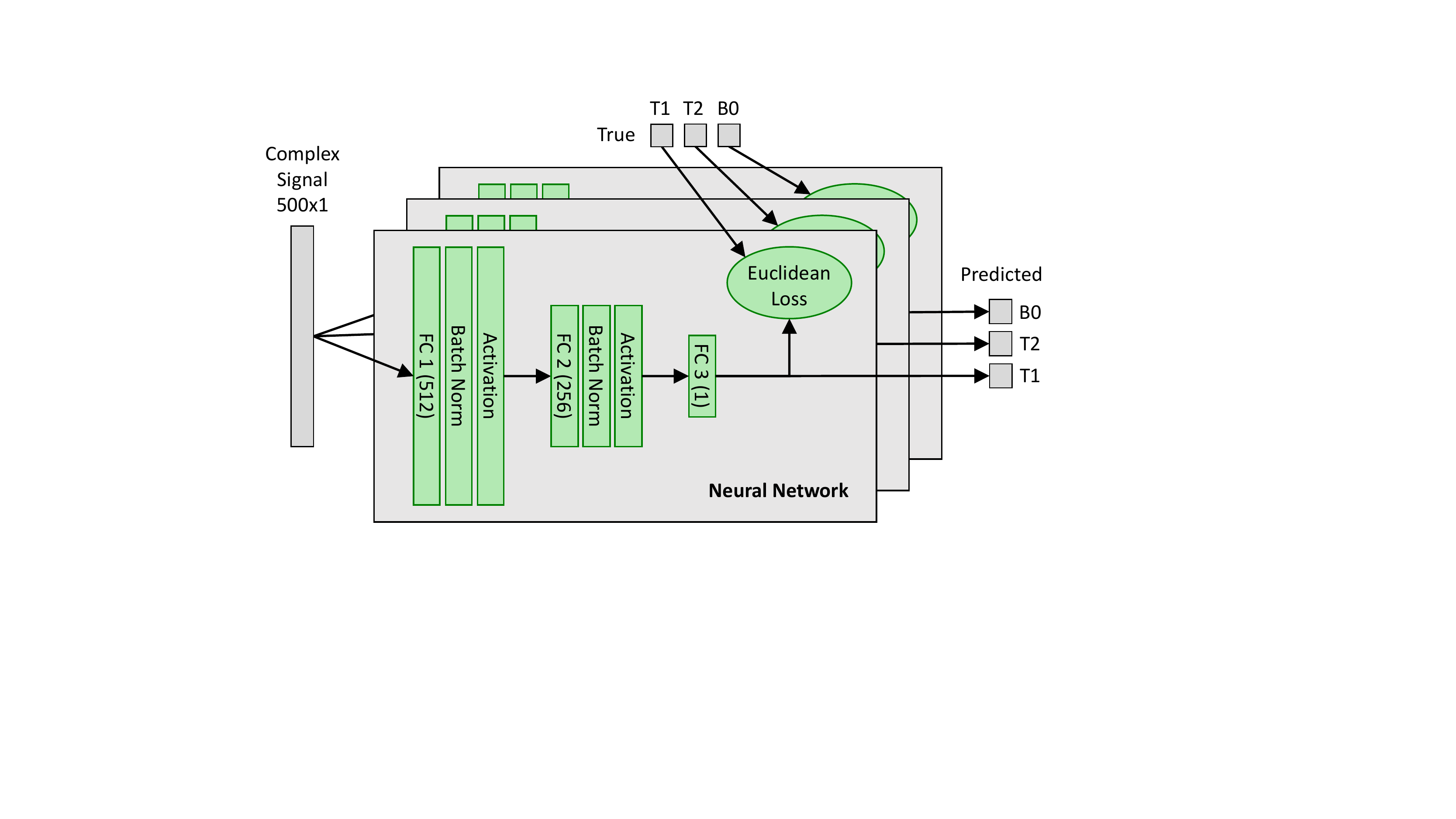}
	\caption{Fully connected neural network architecture, repeated for each desired output label (T1, T2, B0).}
	\label{fig:net}
\end{figure}

{\bf Methods.}  Fig.\ref{fig:net} shows our 3-layer neural network architecture.  We compare six MRI fingerprinting methods.
\begin{enumerate}
\setlength{\itemsep}{0pt}
\item Baseline inner product nearest neighbor and T1, T2, B0 dictionary setup used in \cite{mrf_nature}.
\item Real-valued neural nets with 2-channel real/imaginary inputs representing complex MRI signals, using the ReLU activation function.
\item Real-valued neural nets that are twice as wide as the second model, with 1024 and 512 feature channels in the two hidden layers.
\item Complex-valued neural nets with 1-channel complex MRI signals, using our new cardioid activation function.
\item Complex-valued neural nets using separable sigmoid activation functions (i.e. sigmoid applied to real and imaginary independently) \cite{complex_backprop_nitta}.
\item Complex-valued neural nets using the siglog activation function \cite{complex_backprop_georgiou}.
\end{enumerate}
Here we focus on pixel-wise fingerprinting reconstruction.  We plan to extend our approach to full image predictions for under-sampled MRI fingerprinting in the future.

{\bf Deep Learning Implementation.}
 We implement all our networks in Caffe \cite{caffe}.  For the complex-valued neural nets, we extend the Caffe platform with complex versions of the fully connected layer, batch normalization layer, and complex activation layers, including the $\mathbb{CR}$ calculus back propagation for all the layer functions.



\begin{figure}[b]
\centering
	\includegraphics[width=1.0\linewidth,keepaspectratio]{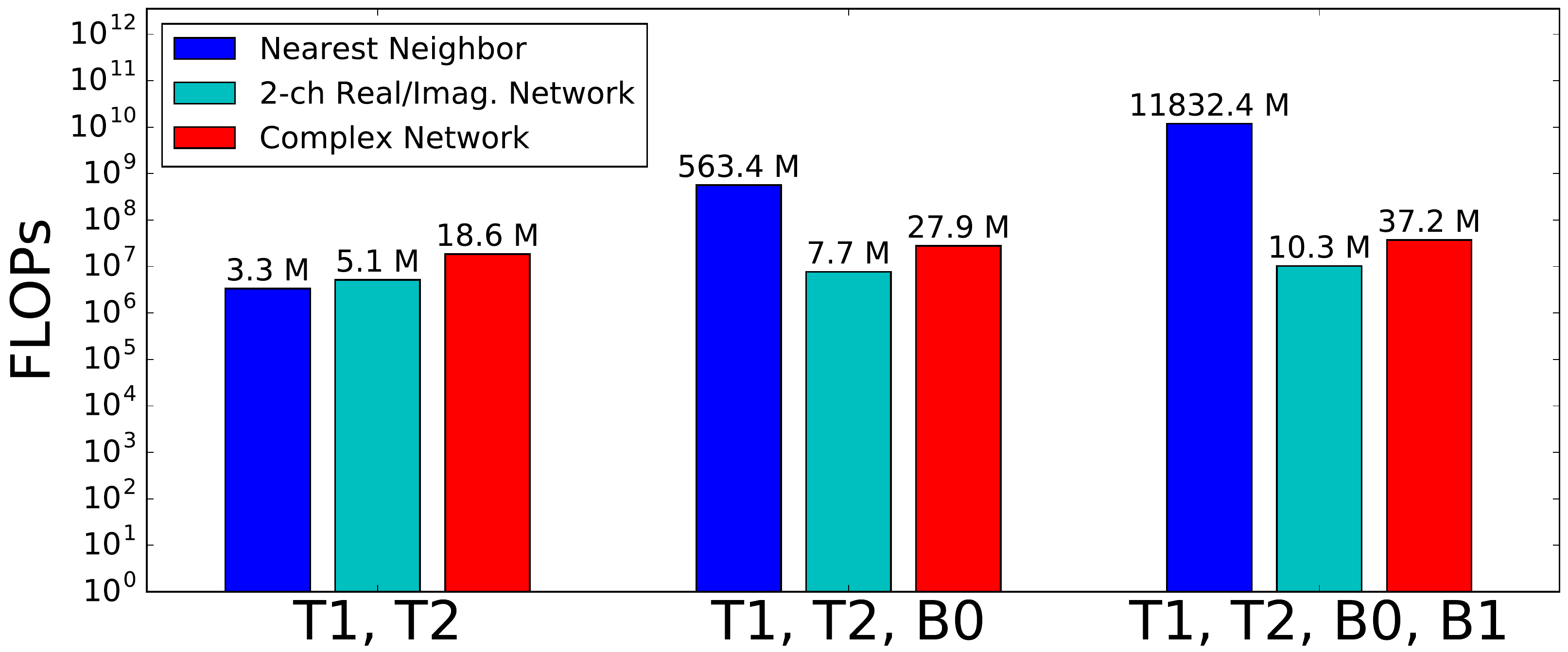}
	\caption{Comparison of floating point operations required to compute the parameters for a single pixel. Note the log scale.}
	\label{fig:flops}
\end{figure}

{\bf Results.} 
Tables \ref{tab:results_clean} and \ref{tab:results_pSNR40} compare the prediction accuracy at no noise and pSNR=40 noise level, respectively.  
Fig. \ref{fig:results_pSNR40} shows sample reconstruction results. Fig. \ref{fig:flops} compares the computational efficiency  in terms of the number of floating point operations (FLOPs). We observe the following:

\begin{table}[t]
\caption{NRMSE results: fingerprinting from clean signals.}
\label{table:results}
\begin{tabular}{p{4.5cm}p{0.8cm}p{0.8cm}p{0.8cm}}
\toprule
\textbf{Network} & \textbf{T1} & \textbf{T2} & \textbf{$\Delta$ B0} \\
\midrule
Nearest neighbor & 10.63 & 39.78 & \textbf{1.02} \\
2-ch real/imaginary network & 2.71 & 8.21 & 2.11 \\
2-ch real/imaginary network 2x & 2.21 & 8.04 & 2.44 \\
\textbf{Complex (cardioid)} & \textbf{1.42} & \textbf{4.34} & 1.32 \\
Complex (separable sigmoid) & 4.72 & 9.24 & 3.33 \\
Complex (siglog) & 2.99 & 12.04 & 3.05 \\
\bottomrule
\end{tabular}
\label{tab:results_clean}
\end{table}

\begin{table}[t]
\caption{NRMSE results: fingerprinting from noisy signals.}
\label{table:results}
\begin{tabular}{p{4.5cm}p{0.8cm}p{0.8cm}p{0.8cm}}
\toprule
\textbf{Network} & \textbf{T1} & \textbf{T2} & \textbf{$\Delta$ B0} \\
\midrule
Nearest neighbor & 12.21 & 40.38 & \textbf{1.08} \\
2-ch real/imaginary network & 11.15 & \textbf{17.96} & 5.23 \\
2-ch real/imaginary network 2x & 11.08 & 22.15 & 7.08 \\
\textbf{Complex (cardioid)} & \textbf{9.40} & 20.98 & 4.43 \\
Complex (separable sigmoid) & 17.31 & 33.09 & 18.83 \\
Complex (siglog) & 102.22 & 237.88 & 266.33 \\
\bottomrule
\end{tabular}
\label{tab:results_pSNR40}
\end{table}

\def\widthA{0.15}
\def\widthB{0.3}

\begin{figure*}[t]
\centering
\tb{@{}ccc@{}}{0.5}{
\toprule
\textbf{T1 (error 5x)} & \textbf{T2 (error 5x)} & \textbf{$\Delta$ B0 (error 5x)}\\
\midrule
\multicolumn{3}{l}{Nearest Neighbor Baseline}\\ [0.75ex]
\tb{cc}{\widthB}{
	\imw{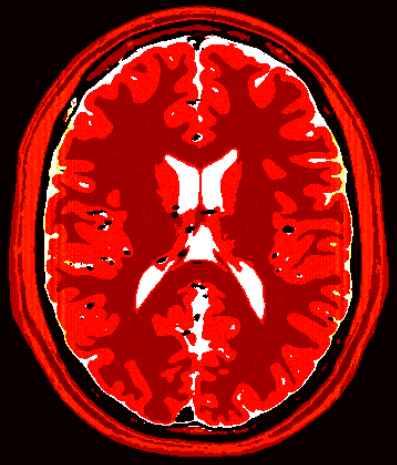}{\widthA} &
	\imw{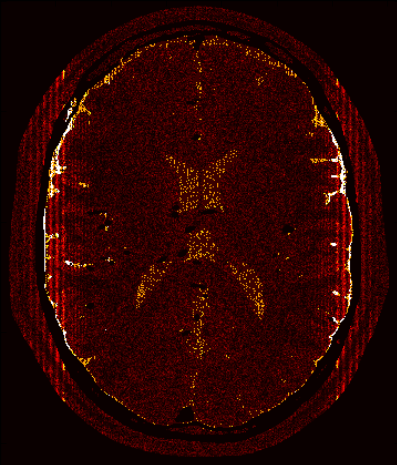}{\widthA}
} &
\tb{cc}{\widthB}{
	\imw{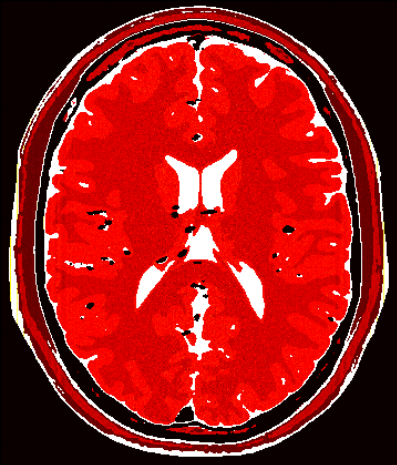}{\widthA} &
	\imw{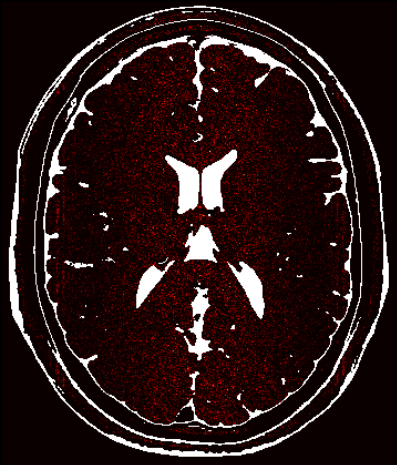}{\widthA}
} &
\tb{cc}{\widthB}{
	\imw{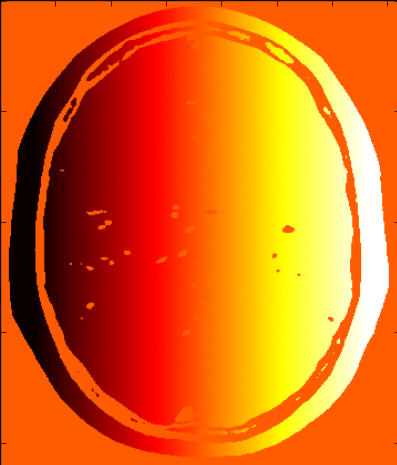}{\widthA} &
	\imw{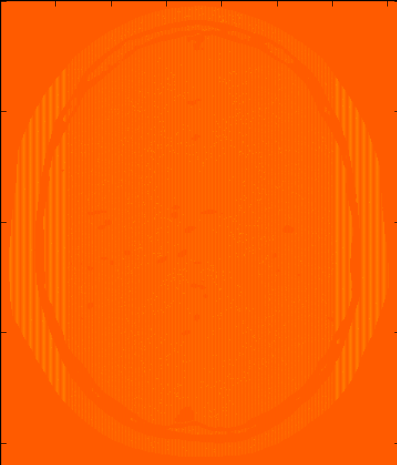}{\widthA}
}
\\
\midrule
\multicolumn{3}{l}{2-ch Real/Imaginary Neural Network}\\ [0.75ex]
\tb{cc}{\widthB}{
	\imw{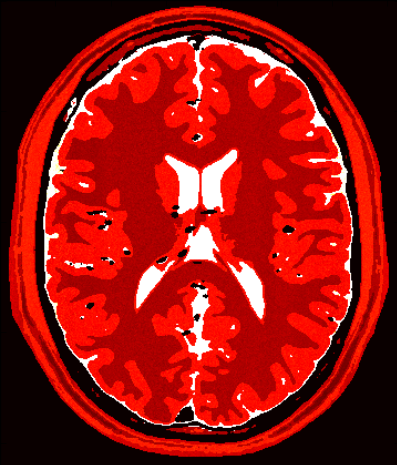}{\widthA} &
	\imw{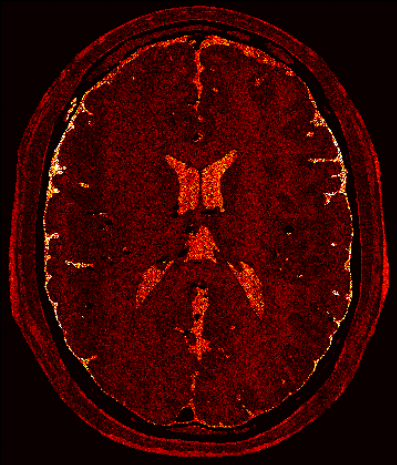}{\widthA}
} &
\tb{cc}{\widthB}{
	\imw{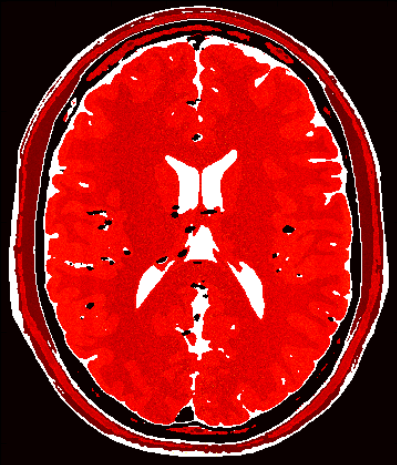}{\widthA} &
	\imw{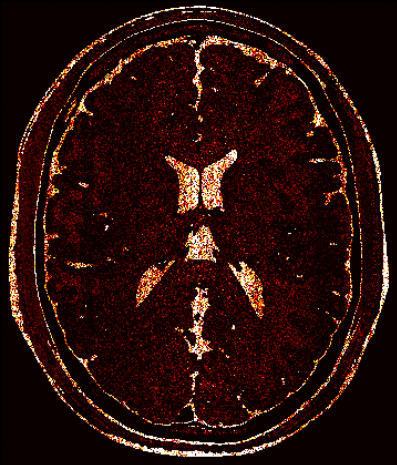}{\widthA}
} &
\tb{cc}{\widthB}{
	\imw{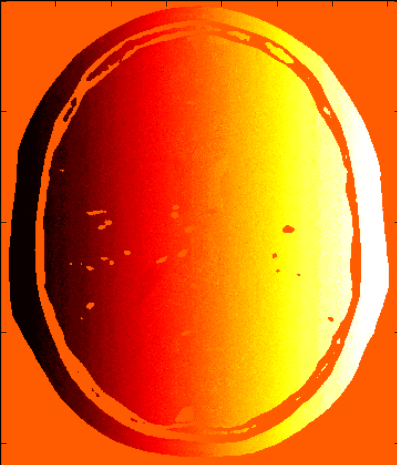}{\widthA} &
	\imw{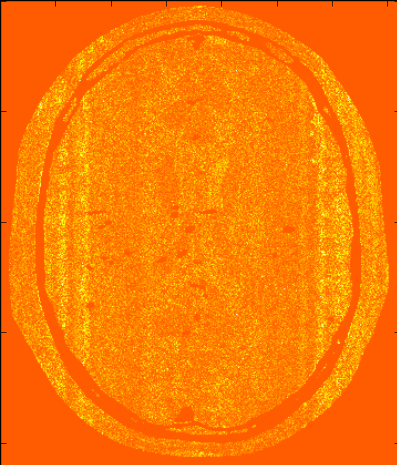}{\widthA}
}
\\
\midrule
\multicolumn{3}{l}{Complex Neural Network (Cardioid)}\\ [0.75ex]
\tb{cc}{\widthB}{
	\imw{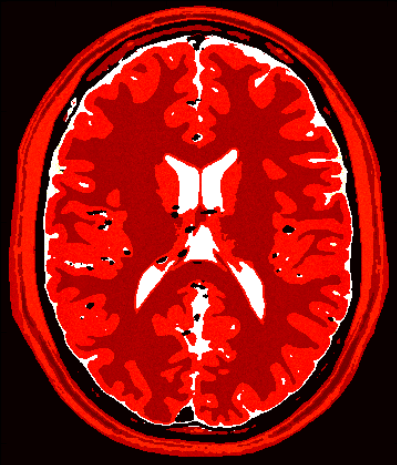}{\widthA} &
	\imw{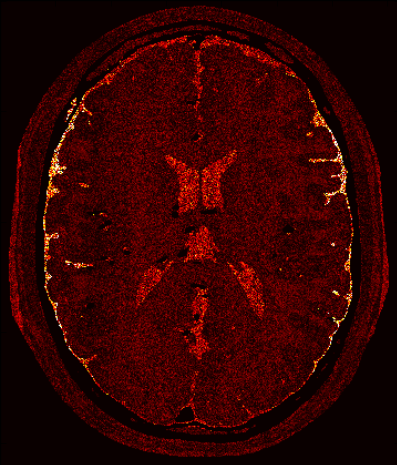}{\widthA}
} &
\tb{cc}{\widthB}{
	\imw{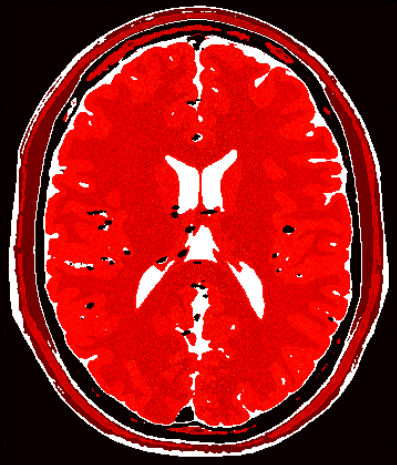}{\widthA} &
	\imw{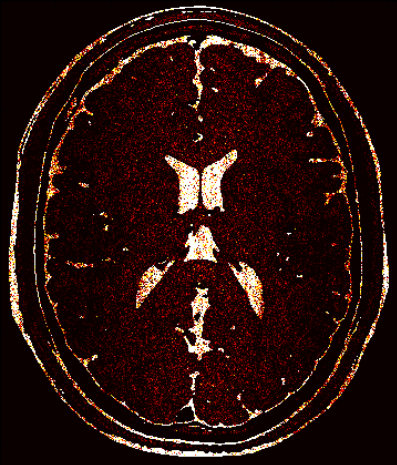}{\widthA}
} &
\tb{cc}{\widthB}{
	\imw{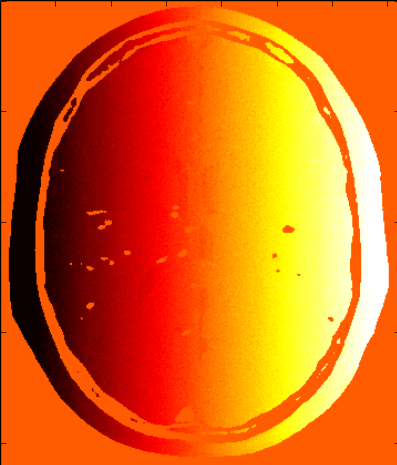}{\widthA} &
	\imw{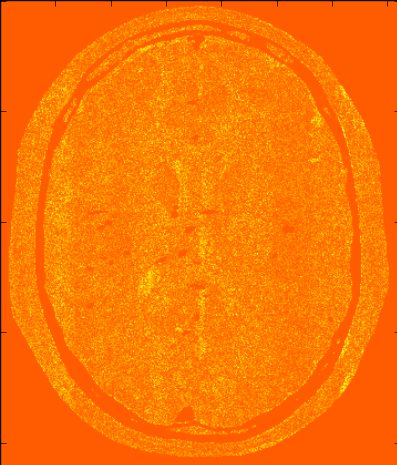}{\widthA}
}
\\
\bottomrule
}
\caption{Numerical phantom with added noise (pSNR=40). Predicted quantitative parameters maps images are shown adjacent to the error image. For visualization purposes, the error images are displayed at 5x the scale of the images. }
\label{fig:results_pSNR40}
\end{figure*}

\begin{enumerate}
\setlength{\itemsep}{0pt}
\item 
A dictionary based approach explodes exponentially with more outputs and becomes infeasible.  
Compared to the two outputs (T1,T2), the \#FLOPs increases by 171$\times$ for the three outputs (T1,T2,B0), 
and by 3,585$\times$ for the four outputs (T1,T2,B0,B1).
\item
Inverse mapping by neural nets outperforms the traditional nearest neighbor baseline on T1 and T2 values, whereas the nearest neighbor approach predicts B0 values more accurately.
\item
Complex-valued neural networks outperform 2-channel real-valued networks for almost all of our experiments, and this advantage cannot be explained away by the twice large model capacity, suggesting that complex-valued networks can bring out information in the complex data more effectively than treating them as arbitrary two-channel real data.
\item
The complex cardioid activation significantly outperformed both the separable sigmoid and siglog activation functions, allowing complex networks to not only compete with, but surpass, real-valued networks.
\end{enumerate}

{\bf Summary.}
For the complex-valued MRI fingerprinting problem, 
we propose a deep learning approach that implements an efficient
nonlinear inverse mapping function that turns MRI signals to tissue parameters directly\footnote{A conference abstract exploring neural networks for MRI fingerprinting \cite{mrf_net_ismrm17} was concurrently published with this work.}.
We generate synthetic (tissue, MRI) data from an MRI simulator, and use
them to train a neural network.  We develop a novel cardioid activation function that enables the successful real-world application of complex-valued neural networks.  Our results demonstrate that complex-valued nets can be more accurate than real-valued nets at complex-valued MRI fingerprinting.  

\section{Acknowledgements}

This research was supported in part by the National Institutes of Health  R01EB009690 grant and a Sloan Research Fellowship. We thank  Michael Kellman, Frank Ong, Jonathan Tamir, and Hong Shang for great discussions about complex calculus, fingerprinting, pulse sequences, and simulator software.

\bibliographystyle{IEEEbib}
\bibliography{refs}

\end{document}

%% file: abstract.tex
The task of MRI fingerprinting is to identify tissue parameters from
complex-valued MRI signals.  The prevalent approach is
dictionary based, where a test MRI signal is compared to stored MRI
signals with known tissue parameters and the most similar
signals and tissue parameters retrieved.  Such an approach does
not scale with the number of parameters and is rather slow when the
tissue parameter space is large.  

Our first novel contribution is to use deep learning as an efficient
nonlinear inverse mapping approach.  We
generate synthetic (tissue, MRI) data from an MRI simulator, and use
them to train a deep net to map the MRI signal to the tissue
parameters directly.
Our second novel contribution is to develop a complex-valued neural network
with new cardioid activation functions.  Our results demonstrate 
that complex-valued neural nets could be much more accurate than real-valued
neural nets at complex-valued MRI fingerprinting.  